\documentclass{article}

\usepackage[nonatbib,preprint]{neurips_2024}

\usepackage[utf8]{inputenc} 
\usepackage[T1]{fontenc}    
\usepackage{hyperref}       
\usepackage{url}            
\usepackage{booktabs}       
\usepackage{amsfonts}       
\usepackage{nicefrac}       
\usepackage{microtype}      
\usepackage{xcolor}         
\usepackage{amsmath,mathtools}
\usepackage{amsthm}
\usepackage{amssymb}
\usepackage{graphicx}
\usepackage[square,numbers]{natbib}
\title{H-SIREN: Improving implicit neural representations \\with hyperbolic periodic functions}

\author{%
  Rui Gao \\
  Department of Mechanical Engineering\\
  The University of British Columbia\\
  Vancouver, BC\\
  \texttt{garrygao@mail.ubc.ca} \\
   \And
 Rajeev K. Jaiman\\
 Department of Mechanical Engineering\\
 The University of British Columbia\\
 Vancouver, BC\\
 \texttt{rjaiman@mail.ubc.ca} \\
}

\begin{document}

\maketitle

\begin{abstract}
	Implicit neural representations (INR) have been recently adopted in various applications ranging from computer vision tasks to physics simulations by solving partial differential equations. Among existing INR-based works, multi-layer perceptrons with sinusoidal activation functions find widespread applications and are also frequently treated as a baseline for the development of better activation functions for INR applications. Recent investigations claim that the use of sinusoidal activation functions could be sub-optimal due to their limited supported frequency set as well as their tendency to generate over-smoothed solutions. We provide a simple solution to mitigate such an issue by changing the activation function at the first layer from $\sin(x)$ to $\sin(\sinh(2x))$. We demonstrate H-SIREN in various computer vision and fluid flow problems, where it surpasses the performance of several state-of-the-art INRs.
\end{abstract}

\section{Introduction}
Implicit neural representations (INR), which maps from a set of coordinates to signals, have been widely adopted to serve as continuous, differentiable representations for various signals, entities or systems. The continuity of INRs means that they are not strictly limited by the data resolution, making it suitable for representing fine details as a function of input coordinates. Implicit neural representations have been applied to a variety of tasks including audio, image, or video representations \cite{chen2024hoin,li2023asmr,lu2023learning,muller2022instant,sitzmann2020implicit}, 3D scene or shape representations \cite{busching2023flowibr,li2024learning,liu2020neural,muller2022instant,park2019deepsdf,sitzmann2020implicit,takikawa2021neural}, etc.

Among existing INR-based works, the sinusoidal activation function \cite{sitzmann2020implicit} (SIREN) see widespread applications. Being able to generate better representations of images, videos, and scenes than ReLU-activated MLPs without the labor of positional encoding, multi-layer perceptron (MLP) with sinusoidal activation function has been frequently adopted as the baseline for the development of better activation functions \cite{liu2023finer,ramasinghe2022beyond,saragadam2023wire,saratchandran2024sampling,shen2023trident} and architectures \cite{chen2024hoin,hao2022implicit,kazerouni2024incode,li2023asmr,li2024learning,lindell2022bacon,martel2021acorn,xie2023diner} for INR applications.

Various recent researchers \cite{liu2023finer,martel2021acorn,saragadam2023wire} have either pointed out or directly demonstrated that whilst an INR with sinusoidal activation function is able to learn smooth representations, the fitted representation could be too smooth in some scenarios. In particular, the fine details are smoothed out in the learned representations. The underlying reason could be that the supported frequency set of sinusoidal activation function depends largely on the tuning of a hyperparameter \cite{yuce2022structured}, and that SIREN can suffer difficulty in representing signals with a wide spectrum of frequencies \cite{liu2023finer}.

A recent effort by Liu et al. \cite{liu2023finer} attempts to address this issue of sinusoidal activation functions by introducing a second-order term such that the outputs of the linear combination after each layer is activated by $\sin(x(1+|x|))$ rather than $\sin(x)$. Since the frequency of the activation function varies for different initialized biases, they also propose to initialize the bias within a larger range instead of the common practice. The modified activation function is shown to reach state-of-the-art performance, surpassing a series of other activation functions in INR applications.

It is possible to show, however, that the inclusion of the second order term will cause the network to gradually bias towards higher frequencies over the network layers, which is not preferable since such bias can lead to overfitting \cite{ramasinghe2023much,tancik2020fourier}. Such overfitting behavior can be demonstrated via a simple experiment, as will be presented later in Sec. \ref{sec:method}. It is therefore preferable to further improve the aforementioned activation function such that it can support a wider range of frequencies whilst still biased towards low frequencies. 

We notice that the design of SIREN largely preserves the supported frequency set over the layers. Therefore, it is possible to adjust the supported frequency set of SIREN by only modifying the activation function at the first layer. For that first layer, we extend from the idea of Liu et al. to infinite orders instead of truncating at second order, eventually leading to the activation function of $\sin(\sinh(2x))$ after some stabilization and simplification efforts. We coin the name H-SIREN for the proposed network as it is an extension of SIREN and uses an additional hyperbolic sinusoidal function in the first layer.

We will demonstrate the effectiveness of the H-SIREN activation function in a series of tasks that include implicit representation of signals. In particular, we report results on image fitting, video fitting, video super-resolution, fitting signed distance functions, neural radiance field, and graph neural network-based fluid flow simulation. Our proposed H-SIREN can be easily incorporated into existing frameworks based on implicit neural representations.
In summary, the key contributions of our work include:
\begin{enumerate}
	\item We introduce a novel hyperbolic periodic activation function to extend the efficacy of the sinusoidal activation function in representing high frequency contents.
        \item The proposed technique allows capturing fine details with minor additional computational overhead and can be easily incorporated into existing SIREN-based INRs.
	\item We demonstrate that H-SIREN surpass the performance of several state-of-the-art activation functions when applied to various implicit neural representation tasks.
\end{enumerate}

\section{Related works}
\subsection{Implicit neural representations}
One of the most well-known applications of INRs is the Neural Radiance Field (NeRF) \cite{mildenhall2021nerf} and its many extensions like \cite{barron2021mip,chan2021pi,park2021nerfies,pumarola2021d,wang2021nerf,yu2021plenoctrees,zhang2020nerf}. Apart from NeRF and other applications discussed in the introduction, recent developments of INRs also extend to other domains like solving partial differential equations \cite{raissi2019physics,raissi2020hidden}, dynamical systems \cite{saratchandran2024sampling}, or robotics \cite{khargonkar2023neuralgrasps}.

Earlier works for implicit neural representations like \cite{atzmon2020sal,park2019deepsdf} adopt ReLU-based networks, but it was soon realized that ReLU-base networks bias towards learning low-frequency contents \cite{rahaman2019spectral}, and therefore can only capture coarse details when used for INRs. A series of solutions were explored in recent works. Apart from the modifications of the activation functions, other efforts include Fourier positional encoding in real \cite{barron2021mip,mildenhall2021nerf,tancik2020fourier} and complex \cite{shen2023trident} domains, hash encoding \cite{muller2022instant, xie2023diner}, multi-level approaches \cite{hao2022implicit,lindell2022bacon,muller2022instant,shabanov2024banf,shekarforoush2022residual}, partitioned/masked approaches \cite{li2024learning,martel2021acorn}, prior embeddings \cite{kazerouni2024incode,song2023piner}, and many more.

\subsection{Activation functions}
Whilst being developed for implicit neural representation tasks, the sinusoidal activation function \cite{sitzmann2020implicit} itself can serve as the activation for other modern neural networks for different applications in the same way as the more traditional activation functions in the ReLU or sigmoid families. Several other works that improve the performance on INRs by only modifying the activation function without assumptions on the network input or other modifications of the network architecture also belong to this category. Examples include the Gaussian activation function \cite{ramasinghe2022beyond} which is more robust to different initializations, the complex Gabor wavelet activation function \cite{saragadam2023wire} that combines the merit of sinusoidal and Gaussian activation functions, as well as the more recent sine cardinal activation function \cite{saratchandran2024sampling} and variable-periodic activation function \cite{liu2023finer} that improve further in performance. H-SIREN also belongs to this category.

\section{Methodology}
\label{sec:method}
In this section, we will start with a brief description of SIREN \cite{sitzmann2020implicit}, followed by its recent variable-periodic extension \cite{liu2023finer}. We then extend from there and derive step-by-step towards the proposed activation layer, and eventually discuss the concepts of controlling frequency growth.

\paragraph{Sinusoidal activation function}
We start with the SIREN activation function, which for the output from the $(l-1)$-th layer $\boldsymbol{z}^{l-1}$, specifies that
\begin{equation}
	\begin{aligned}
	{x}^{l-1}=&{W}^l{z}^{l-1}+{b}^l\\
	{z}^{l}=&\sin(\omega_0{x}^{l-1})
	\end{aligned}
\end{equation}
with $\omega_0=30$. The weights are initialized by
\begin{equation}
	\begin{aligned}
		{W}^l&\sim\mathcal{U}(-1/n,1/n)\quad\quad\quad\quad\quad l=1\\
		{W}^l&\sim\mathcal{U}(-\sqrt{6/n}/\omega_0,\sqrt{6/n}/\omega_0)\quad\quad l=2,3,\ldots,n_l
	\end{aligned}
\end{equation}
for a multi-layer perceptron with $n_l$ hidden layers, where $n$ denotes the number of input features to the layer, and $\mathcal{U}$ denotes uniform distribution.

\paragraph{Variable periodic extension}
Recently, Liu et al. \cite{liu2023finer} proposed to expand the supported frequency set of SIREN by introducing a variable-periodic extension, with which the output from the $(l-1)$-th layer $\boldsymbol{z}^{l-1}$ goes through
\begin{equation}
	\begin{aligned}
	{x}^{l-1}=&{W}^l{z}^{l-1}+{b}^l\\
	{z}^{l}=&\sin(\omega_0{x}^{l-1}(1+|{x}^{l-1}|))
	\end{aligned}
\end{equation}
with the same default $\omega_0$ and the weight initialization as SIREN while a different bias initialization 
\begin{equation}
	{b}\sim\mathcal{U}(-k,k),
\end{equation}
where $k$ varies for different problems. The inclusion of second-order term $x^2$ and the greater bias initialization range helps to fit signals and features of higher frequencies.

\begin{figure}
	\centering
	\includegraphics[]{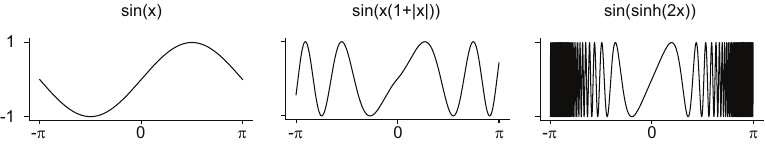}
	\caption{Plots comparing different activation functions: SIREN (left), FINER (middle) and H-SIREN (right).}
	\label{fig:activations}
\end{figure}

\paragraph{Extending to infinite orders}
If adding a second order $x^2$ within the $\sin$ function helps learning higher frequency information, it is intuitive for one to hypothesize that extending to infinite orders might also be beneficial. Writing $x(1+|x|)$ as its alternative form $x+\operatorname{sgn}(x)x^2$, we can treat it as the second order truncation of the sum of an infinite series, for which the untruncated form would be
\begin{equation}
\label{eq:inforder}
	\begin{aligned}
		z=&\sin(\omega_0f(x))\\
		f(x)&=x+\operatorname{sgn}(x)x^2+x^3+\operatorname{sgn}(x)x^4+\cdots\\
		&=\frac{x}{1-\operatorname{sgn}(x)x}\\
		&=\operatorname{softsign}^{-1}(x)
	\end{aligned}
\end{equation}
for $x\in(-1,1)$, where $\operatorname{sgn}(\cdot)$ denotes the sign function. Two singularity points lie in $x=\pm1$, which limits us to either initialize the weights and biases such that the values of $x=Wz+b$ are strictly within $(-1,1)$, or using a function to limit its value before sending to the inverse of softsign function. The latter option is obviously easier to implement. Two common functions used to limit the value of $f(x)$ to the range of $(-1,1)$ for any input $x$ are the softsign function itself and the hyperbolic tangent function, meaning that the hyperbolic tangent function is the only remaining candidate for this case. Replacing $x$ by $\tanh(x)$ in Eq. \ref{eq:inforder} gives us
\begin{equation}
	\label{eq:sinh}
	\begin{aligned}
		f(x)&=\operatorname{softsign}^{-1}(\tanh(x))\\
		&=\sinh(x)\exp(|x|)\\
		&\approx\sinh(2x)
	\end{aligned}
\end{equation}
The approximation in the last line reduces the computational cost and smoothens the second-order derivative. We illustrate the proposed activation function versus SIREN and FINER in Fig. \ref{fig:activations}.

\paragraph{Controlling frequency growth}
With the original SIREN activation function, the frequency spectrum at initialization is largely consistent across layers, and the maximum frequency only grows slowly over the layers. When we extend to orders higher than one, however, it is noticed that the spectrum gradually evolves towards higher frequency with the increase of layer number. We demonstrate this trend in Fig. \ref{fig:spec}a, details in generating the spectrum are included in Appendix \ref{sec:appspec}. Such an increase in bias towards higher frequencies will eventually cause the network to overfit when used to learn low frequency representations. To demonstrate this, we construct a mini-experiment, where SIREN and FINER with 5 hidden layers of layer width 64 are used to fit the function $y=f(x)=\frac{\sin(30x)}{10x}$. A total of 200 points are sampled with equal distance within $x\in[-1,1]$, and the fitted function after 1000 iterations of gradient descent is evaluated at 2000 equidistant points within $x\in[-1,1]$. The results are shown in Fig. \ref{fig:spec}b. It is clear that the bias towards high frequencies cause severe overfitting. In order to control the frequency increase, we only apply Eq. \ref{eq:sinh} to the first layer of the network, whilst keeping the sinusoidal activation function for other layers. The resulting network preserves the frequency distribution over the layers, and supports a large range of frequencies. It is also biased towards low frequencies, which makes it more robust to overfitting, as demonstrated in Fig. \ref{fig:spec}b.

\begin{figure}
	\centering
	\includegraphics[]{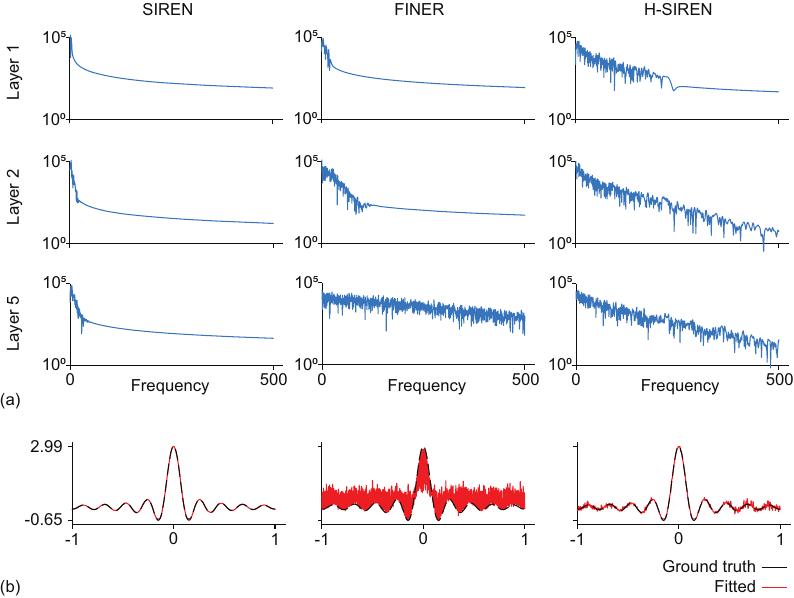}
	\caption{(a) Activation spectrum after hidden layer 1, 2 and 5, for MLP with different activation functions at initialization. (b) Results for fitting a simple function using MLPs with different activation functions.}
	\label{fig:spec}
\end{figure}

\section{Experiments}
\label{sec:exp}
We verify the performance of the proposed activation function against the baseline SIREN, the recently proposed FINER, as well as WIRE, another recently-proposed activation function that combines the merit of sinusoidal and Gaussian activation functions via the use of complex Gabor wavelets. We report four different computer vision test cases, namely fitting images ($\mathbb{R}^2\rightarrow\mathbb{R}^3$), representing videos ($\mathbb{R}^3\rightarrow\mathbb{R}^3$) with its side task of video super-resolution ($\mathbb{R}^3\rightarrow\mathbb{R}^3$), and novel view synthesis with NeRF ($\mathbb{R}^5\rightarrow\mathbb{R}^4$). We additionally report the results for representing signed distance fields ($\mathbb{R}^3\rightarrow\mathbb{R}$) in the appendix. Besides computer vision tasks, we also report the results when the INRs are used for learning the evolution of fluid flow systems under a graph neural network framework ($\mathcal{G}\rightarrow\mathcal{G}$).

\paragraph{Implementation} 
We use the open-source implementations from the respective authors of SIREN and WIRE (both with MIT licence), and implemented FINER following the descriptions in \cite{liu2023finer}. For SIREN, we take $\omega_0=30$ throughout. For WIRE, different values of $\omega_0$ and $s_0$ are used for different tasks, details are reported for each experiment in their respective subsections and respective appendix sections. For FINER \cite{liu2023finer}, since the authors did not include the schemes to tune the value of $k$ or a default value of $k$, we use their $k$ value if the same task is tested in their work, and test over different possible $k\in\{0.6,0.7,0.8,0.9,1.0,\sqrt{1/n}\}$ for other cases, within which the last one is the default bias initialization in PyTorch. All networks are trained with Adam optimizer \cite{kingma2014adam} with $\beta_1=0.9$ and $\beta_2=0.999$. All models are implemented with PyTorch \cite{paszke2019pytorch}, and executed on a single Nvidia RTX 3090 GPU with CPU being AMD Ryzen 9 5900 @ 3.00 GHz. Other implementation details specific to each of the experiments will be described in their respective subsection and their respective appendix sections. All code and data will be made publicly available.

\paragraph{Training and inference speed} We report the training and inference speed for all the experiments in appendix \ref{sec:appspeed}. Networks with H-SIREN is only slightly slower than SIREN due to the additional computational steps, but faster than FINER and WIRE.

\subsection{Fitting 2D images}
\begin{figure}
	\centering
	\includegraphics[]{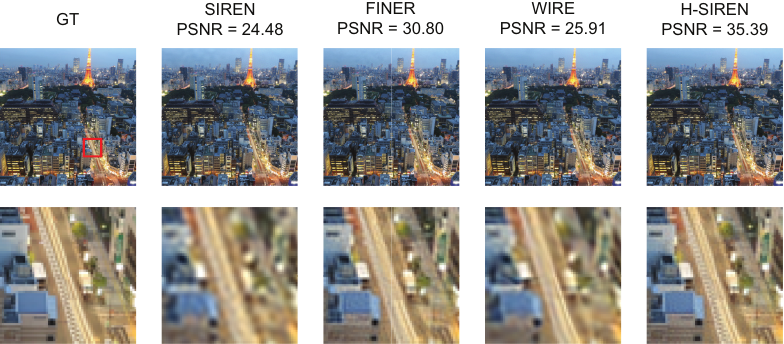}
	\caption{Comparison between INRs with different activation functions for fitting 2D images.}
	\label{fig:fitimage}
\end{figure}

We choose simple 2D image fitting as the first task, and the "natural" data set \cite{tancik2020fourier} is selected for this purpose. Specifically, we use the training partition of this data set that contains 16 figures. We report the main results with all the networks being implemented with two hidden layers of width 512. Some other combinations of layer width and layer depth are also tested, and those results are attached in Appendix \ref{sec:appfitimage}. For FINER, we use $k=\frac{\sqrt{2}}{2}$ following the choice in \cite{liu2023finer}. For WIRE, we use $\omega_0=20$ and $\sigma_0=10$ following the choice in \cite{saragadam2023wire}. All the networks are trained with gradient descent (i.e., batch size equal to the image size) for 5000 iterations. 
We demonstrate the results in Fig. \ref{fig:fitimage}. H-SIREN is the only one able to clearly capture the road traffic. We additionally report the peak signal-to-noise ratio (PSNR) and structural similarity index measure (SSIM) values for the whole data set in Table \ref{tab:fitimage}.

\begin{table}
	\caption{Quantitative measures of the results for image fitting cases.}
	\label{tab:fitimage}
	\centering
	\begin{tabular}{lllll}
		\toprule
		Method & SIREN & FINER & WIRE & H-SIREN \\
		\midrule
		PSNR & $26.87\pm3.18$ & $33.86\pm3.36$ & $28.51\pm3.15$& $\mathbf{37.80\pm3.10}$ \\
		SSIM & $0.808\pm0.065$ & $0.948\pm0.020$ & $0.853\pm0.061$ & $\mathbf{0.973\pm0.010}$ \\
		\bottomrule
	\end{tabular}
\end{table}

\subsection{Representing video}
For the video representation task, we consider three different videos: The "cat" and "bike" video that were previously tested by Sitzmann et al. \cite{sitzmann2020implicit}, and also a new "bush" video that we collect and process ourselves, which features some shrubs shaking in the wind. The INRs are implemented with three hidden layers of width 1024 for all three videos, and trained for 150 epochs. The Pytorch default initialization of bias (i.e., $k=\sqrt{1/n}$) is used for FINER since it gives the best result among the tested options. For WIRE, we use the same $\omega_0=20$ and $\sigma_0=10$ as those used in image fitting tasks.

\begin{figure}
	\centering
	\includegraphics[]{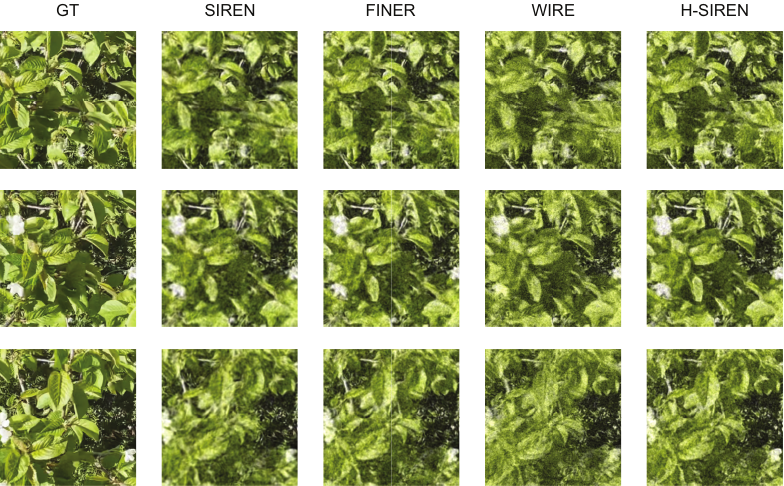}
	\caption{Comparison among INRs with different activation functions for fitting the "bush" video. From top to bottom row: frame 10, 200, 290.}
	\label{fig:fitvideo}
\end{figure}

We again compare the results by PSNR and SSIM, listed in Table \ref{tab:fitvideo}. For the easier cases of cat and bike, we observe that the proposed network is slightly better in PSNR metric but slightly lower in SSIM than those of FINER and SIREN. For the bush video, which is significantly more difficult to fit, we observe a more pronounced difference between the results of different networks, with H-SIREN having superior performance than INR with other activation functions. We plot out the comparison between the results at several different frames for the bush video in Fig. \ref{fig:fitvideo}.

\begin{table}
	\caption{Quantitative measures of the results for different video fitting and video super-resolution cases. SR: super-resolution.}
	\label{tab:fitvideo}
	\centering
	\begin{tabular}{lllll}
		\toprule
		Video: Cat           &&&&        \\
		\cmidrule(r){1-1}
		Method & SIREN & FINER & WIRE & H-SIREN \\
		\midrule
		PSNR & $29.98\pm1.02$ & $30.48\pm0.95$ & $30.61\pm0.68$& $\mathbf{30.77\pm0.82}$ \\
		SSIM & $0.821\pm0.018$ & $\mathbf{0.822\pm0.014}$ & $0.807\pm0.014$ & $0.817\pm0.012$ \\
		\midrule
		Video: Bike           &&&&        \\
		\cmidrule(r){1-1}
		Method & SIREN & FINER & WIRE & H-SIREN \\
		\midrule
		PSNR & $31.72\pm2.22$ & $32.19\pm1.89$ & $29.77\pm1.35$& $\mathbf{32.24\pm1.66}$ \\
		SSIM & $0.894\pm0.037$ & $\mathbf{0.896\pm0.027}$ & $0.786\pm0.020$ & $0.891\pm0.021$ \\
		\midrule
		Video: Bush           &&&&        \\
		\cmidrule(r){1-1}
		Method & SIREN & FINER & WIRE & H-SIREN \\
		\midrule
		PSNR & $18.95\pm0.64$ & $19.84\pm0.48$ & $19.50\pm0.59$& $\mathbf{20.31\pm0.46}$ \\
		SSIM & $0.488\pm0.042$ & $0.541\pm0.030$ & $0.519\pm0.030$ & $\mathbf{0.568\pm0.025}$ \\
		\midrule
		Video SR: Cat           &&&&        \\
		\cmidrule(r){1-1}
		Method & SIREN & FINER & WIRE & H-SIREN \\
		\midrule
		PSNR & $28.36\pm1.13$ & $28.53\pm1.11$ & $28.17\pm0.75$& $\mathbf{29.08\pm0.88}$ \\
		SSIM & $\mathbf{0.790\pm0.020}$ & $0.780\pm0.015$ & $0.731\pm0.015$ & $0.782\pm0.012$ \\
		\midrule
		Video SR: Bush           &&&&        \\
		\cmidrule(r){1-1}
		Method & SIREN & FINER & WIRE & H-SIREN \\
		\midrule
		PSNR & $16.92\pm0.44$ & $17.51\pm0.40$ & $17.77\pm0.38$& $\mathbf{17.82\pm0.392}$ \\
		SSIM & $0.330\pm0.032$ & $0.374\pm0.030$ & $0.396\pm0.025$ & $\mathbf{0.399\pm0.028}$ \\
		\bottomrule
	\end{tabular}
\end{table}

\paragraph{Side task: Video super-resolution}
As a side task, we also tested the capability of the networks on video super-resolution. We only tested on the "cat" and "bush" videos. For both of these videos, we subsample every frame by every next pixel in both spatial directions. The networks trained on these subsampled videos are tested for their ability to reproduce the original video. We list the statistics in Table \ref{tab:fitvideo}. A few different frames of the "cat" video are plotted in Fig. \ref{fig:videosr} in the appendix for the readers' reference.

\subsection{Application to neural radiance fields}
\begin{figure}
	\centering
	\includegraphics[]{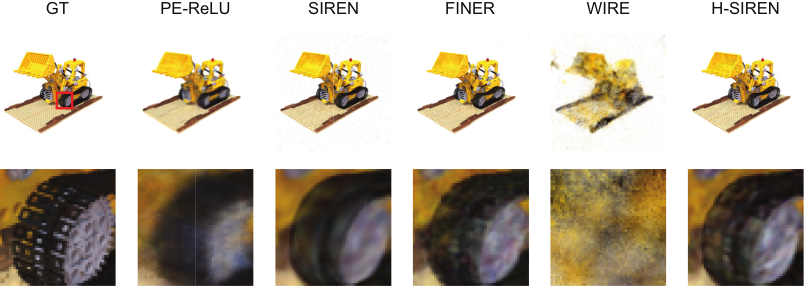}
	\caption{Comparison among different activation functions for NeRF. Test image 62 of the "lego" scene is plotted.}
	\label{fig:nerf}
\end{figure}

One of the most well-known applications of INRs is the so-called neural radiance field (NeRF) \cite{mildenhall2021nerf} for novel view synthesis, which learns a map from location ($\mathbb{R}^3$) and view angle ($\mathbb{R}^2$) to color ($\mathbb{R}^3$) and opacity ($\mathbb{R}$). The images from different views are generated from this learned map via differentiable volume rendering \cite{kajiya1984ray}. The original NeRF architecture used Rectified Linear Units as an activation function and therefore relies on positional encoding, but it has been shown \cite{liu2023finer,saragadam2023wire} that INRs with some modern activation functions do not necessarily need such treatment. We therefore remove the positional encodings in the tests. We also performed some other modifications to the original NeRF implementation like reducing the number of layers, details are attached in Appendix \ref{sec:appnerf}. We test on half of the Blender data set \cite{mildenhall2021nerf}, in particular the "drums", "mic", "lego", and "ship" scenes. We downsample all the images to $400\times400$ in training and tests to avoid memory overflow, and train all the networks for all the cases for 50000 iterations. The qualitative results are shown in Table \ref{tab:nerf} and Fig. \ref{fig:nerf} respectively. H-SIREN gives the best results in three of the four test scenes, and only slightly worse than FINER in the "ship" scene. We also plot out results for test image 62 in "lego" scene for different INRs in Fig. \ref{fig:nerf} for some qualitative comparison.

\begin{table}
	\caption{Quantitative measures of the results for novel view synthesis with NeRF. The results with the postional encoding + ReLU-based INR, used in the original NeRF paper, is also attached for readers' reference.}
	\label{tab:nerf}
	\centering
	\begin{tabular}{llllll}
		\toprule
		Drums           &&&&&        \\
		\cmidrule(r){1-1}
		Method & PE-ReLU & SIREN & FINER & WIRE & H-SIREN \\
		\midrule
		PSNR & $20.39\pm0.68$ & $21.93\pm2.41$ & $22.65\pm1.35$ & $15.06\pm0.78$& $\mathbf{22.72\pm1.46}$ \\
		SSIM & $0.824\pm0.014$ & $0.824\pm0.039$ & $0.855\pm0.021$ & $0.564\pm0.057$ & $\mathbf{0.867\pm0.021}$ \\
		\midrule
		Mic           &&&&&        \\
		\cmidrule(r){1-1}
		Method & PE-ReLU & SIREN & FINER & WIRE & H-SIREN \\
		\midrule
		PSNR & $24.66\pm0.97$ & $27.35\pm3.745$ & $29.11\pm1.40$ & $15.81\pm0.76$& $\mathbf{29.49\pm1.47}$ \\
		SSIM & $0.938\pm0.005$ & $0.945\pm0.019$ & $0.957\pm0.006$ & $0.440\pm0.026$ & $\mathbf{0.960\pm0.005}$ \\
		\midrule
		Lego           &&&&&        \\
		\cmidrule(r){1-1}
		Method & PE-ReLU & SIREN & FINER & WIRE & H-SIREN \\
		\midrule
		PSNR & $23.57\pm0.95$ & $24.88\pm2.68$ & $26.40\pm1.27$ & $16.85\pm1.26$ & $\mathbf{26.75\pm1.35}$ \\
		SSIM & $0.827\pm0.022$ & $0.859\pm0.023$ & $0.893\pm0.016$ & $0.598\pm0.061$ & $\mathbf{0.900\pm0.017}$ \\
		\midrule
		Ship           &&&&&        \\
		\cmidrule(r){1-1}
		Method & PE-ReLU & SIREN & FINER & WIRE & H-SIREN \\
		\midrule
		PSNR & $23.00\pm1.17$ & $23.61\pm4.35$ & $\mathbf{24.93\pm1.96}$ & $16.12\pm2.42$ & $24.91\pm2.20$ \\
		SSIM & $0.761\pm0.043$ & $0.774\pm0.051$ & $\mathbf{0.785\pm0.036}$ & $0.434\pm0.039$ & $0.782\pm0.036$ \\
		\bottomrule
	\end{tabular}
\end{table}

\subsection{Graph neural network-based fluid flow simulation}
\begin{figure}
	\centering
	\includegraphics[]{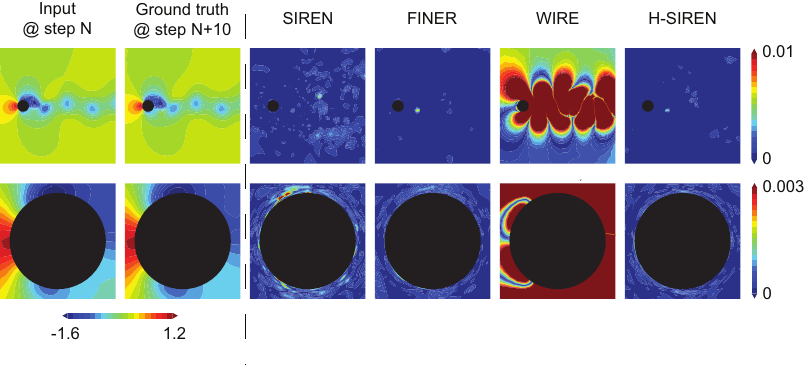}
	\caption{Comparison among different methods for GNN-based fluid flow prediction. The ground truth input non-dimensionalized pressure field and the ground truth field to be predicted are plotted in the leftmost two columns. Absolute prediction error fields are plotted for different activation functions instead of the predicted pressure field itself to facilitate visual comparison.}
	\label{fig:gnn}
\end{figure}

Multi-layer perceptrons with sinusoidal activation function were recently introduced \cite{gao2024finite} as the update functions for graph neural network-based fluid flow simulations, and demonstrate an advantage over ReLU-based update functions. It is therefore of interest to examine the performance of different INRs for such a purpose. We perform such tests on a flow around cylinder data set and a flow around NACA0012 airfoil data set, with details described in the Appendix \ref{sec:appgnn}. We adopt the MeshGraphNet \cite{pfaff2020learning} architecture, within which we replace the activation function of all the multi-layer perceptrons with the activation function being tested, and reduce the layer depth to 5 since the data set is small. Tests with larger data sets would usually require the employment of training noise schemes, and therefore we refrain from discussing those applications. The results for 1-step and 10-step RMSE are reported in Table \ref{tab:gnn}, and the 10-step prediction error of the pressure field for one sample in the test data set for flow around airfoil are plotted in Fig. \ref{fig:gnn}.

\begin{table}
	\caption{Root-mean-square error of the results for GNN-based fluid flow simulation.}
	\label{tab:gnn}
	\centering
	\begin{tabular}{lllll}
		\toprule
		1-step ($\times10^{-5}$)           &&&&        \\
		\cmidrule(r){1-1}
		Case & SIREN & FINER & WIRE & H-SIREN \\
		\midrule
		Cylinder & $13.1\pm8.1$ & $8.7\pm2.2$ & $807\pm23$& $\mathbf{6.4\pm1.1}$ \\
		Airfoil & $21.2\pm8.4$ & $21.7\pm12.1$ & $980\pm87$ & $\mathbf{20.0\pm9.96}$ \\
		\midrule
		10-step ($\times10^{-4}$)           &&&&        \\
		\cmidrule(r){1-1}
		Case & SIREN & FINER & WIRE & H-SIREN \\
		\midrule
		Cylinder & $9.3\pm5.4$ & $5.7\pm1.5$ & $784\pm23$& $\mathbf{4.1\pm0.6}$ \\
		Airfoil & $15.4\pm5.2$ & $15.0\pm7.5$ & $928\pm78$ & $\mathbf{13.7\pm6.1}$ \\
		\bottomrule
	\end{tabular}
\end{table}

\section{Conclusion}
In this paper, we proposed and assessed the H-SIREN that utilizes the hyperbolic periodic function and captures finer details at a small additional computational cost. 
We demonstrated through a function fitting experiment that H-SIREN supports a wide span of frequencies while being relatively robust to overfitting. 
H-SIREN consistently demonstrated better performance over several other state-of-the-art activation functions in various experiments including image fitting, video representations, video super-resolution, representing signed distance field, learning neural radiance fields, and graph neural network-based fluid flow simulations.

\newpage 
\bibliography{references}


\newpage

\appendix

\section{Generating activation spectrum}
\label{sec:appspec}
The activation spectrum plotted in Fig. \ref{fig:spec}a is similar to that in Sitzmann et al. \cite{sitzmann2020implicit}. The code is acquired from the open-source repository shared by Sitzmann et al. \cite{sitzmann2020implicit}. We use SIREN, FINER, and H-SIREN each with a multi-layer perceptron of five hidden layers and layer width 2048. A total of 10000 equidistant inputs are drawn within $[-1,1]$, and then 1D fast fourier transform is performed on the output after each layer. 

\section{Training and inference speed}
\label{sec:appspeed}
We report the computational time required for the experiments reported in Sec. \ref{sec:exp} in Table \ref{tab:speed}. For these tests, we downclock the CPU and GPU to avoid performance perturbations. The CPU is limited to 2.91 GHz, slightly lower than its base frequency, whilst the GPU is downclocked by 602 MHz.
\begin{table}
	\caption{Training and inference speed for video fitting tasks.}
	\label{tab:speed}
	\centering
	\begin{tabular}{lllll}
		\toprule
		Image           &&&&        \\
		\cmidrule(r){1-1}
		Method & SIREN & FINER & WIRE & H-SIREN \\
		\midrule
		Training & 272s & 404s & 780s & 307s \\
		Inference & <1s & <1s & <1s & <1s \\
		\midrule
		Video: Cat           &&&&        \\
		\cmidrule(r){1-1}
		Method & SIREN & FINER & WIRE & H-SIREN \\
		\midrule
		Training & $\approx$98s/epoch & $\approx$117s/epoch & $\approx$256s/epoch & $\approx$108s/epoch \\
		Inference & 34s & 42s & 83s & 36s \\
		\midrule
		Video SR: Cat            &&&&        \\
		\cmidrule(r){1-1}
		Method & SIREN & FINER & WIRE & H-SIREN \\
		\midrule
		Training & $\approx$61s/epoch & $\approx$74s/epoch & $\approx$147s/epoch & $\approx$65s/epoch \\
		Inference & 35s & 44s & 83s & 38s \\
		\midrule
		NERF: Drums            &&&&        \\
		\cmidrule(r){1-1}
		Method & SIREN & FINER & WIRE & H-SIREN \\
		\midrule
		Training & $\approx$110min & $\approx$147min & $\approx$422min & $\approx$117min \\
		Inference & $\approx$7.0s/image & $\approx$9.5s/image & $\approx$26.9s/image & $\approx$7.7s/image \\
		\midrule
		GNN: cylinder           &&&&        \\
		\cmidrule(r){1-1}
		Method & SIREN & FINER & WIRE & H-SIREN \\
		\midrule
		Training & 301s & 394s & 792s & 322s \\
		Inference & $\approx$7.6ms/step & $\approx$9.7ms/step & $\approx$12.2ms/step & $\approx$8.1ms/step \\
		\bottomrule
	\end{tabular}
\end{table}

\section{Additional experiment: Representing signed distance fields}
\label{sec:appsdf}
For the representation of 3D shapes with signed distance fields, we replicate the experiments by Sitzmann et al. \cite{sitzmann2020implicit} on a Thai statue which available from the Stanford 3D Scanning Repository (\url{https://graphics.stanford.edu/data/3Dscanrep/}, free for use in research purposes). The code for the loss function is acquired from the open source code shared by Sitzmann et al. with MIT licence. For the statue case, the networks are implemented with three hidden layers of width 512, with the exception of FINER which was trained with layer width 513 instead of 512 to avoid a wierd bug.  All random seeds are initialized as 123456789 for all experiments.

\begin{figure}
	\centering
	\includegraphics[]{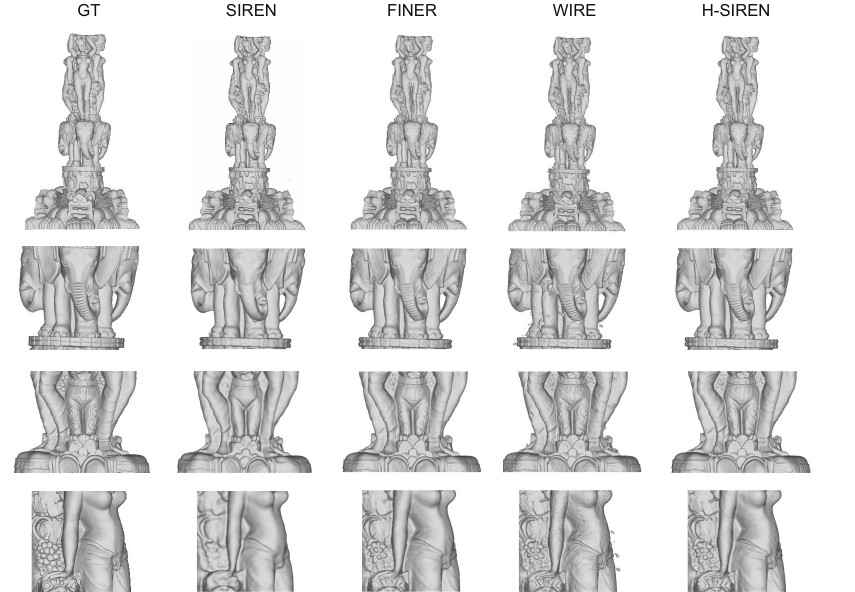}
	\caption{Comparison between different methods for representing the Thai statue.}
	\label{fig:fitsdfthai}
\end{figure}

For the Thai statue case, all the networks are trained for 600 epochs with constant learning rate of $1\times10^{-4}$. The effective batch size is 250000, implemented by choosing a batch size of 50000 and accumulating the gradient over every 5 iterations. Gradient is clipped immediately before optimizer step with maximum norm 1. For FINER, we use $k=1$ following the choice of the authors on signed distance field representation tasks. For H-SIREN, we manually select $r=3$ rather than the default $r=2$ as the statue contains many fine details. The results are plotted in Fig. \ref{fig:fitsdfthai}. We notice that the H-SIREN is able to fit many of the fine details of the statue, as shown in the second and third row of Fig. \ref{fig:fitsdfthai}, and performs better than SIREN or FINER. In the mean time, it also overfits significantly less than WIRE on smooth regions, as shown in the bottom row of Fig. \ref{fig:fitsdfthai}. 

\section{2D image fitting}
\label{sec:appfitimage}
\paragraph{Implementation details} We use the training partition of the publicly available "natural" data set shared by Tancik et al. \cite{tancik2020fourier} with MIT licence, which contains 16 images. Each of the images are trained and evaluated individually. The training for each image last for 5000 iterations with gradient descent at fixed learning rate $10^{-4}$, using mean-squared error (MSE) loss. All random seeds are initialized as 1 for all experiments. All inputs and outputs are normalized to have minimum -1 and maximum 1. All the networks contain two hidden layers with layer width 512 for the results reported in Sec. \ref{sec:exp}. 

\begin{figure}
	\centering
	\includegraphics[]{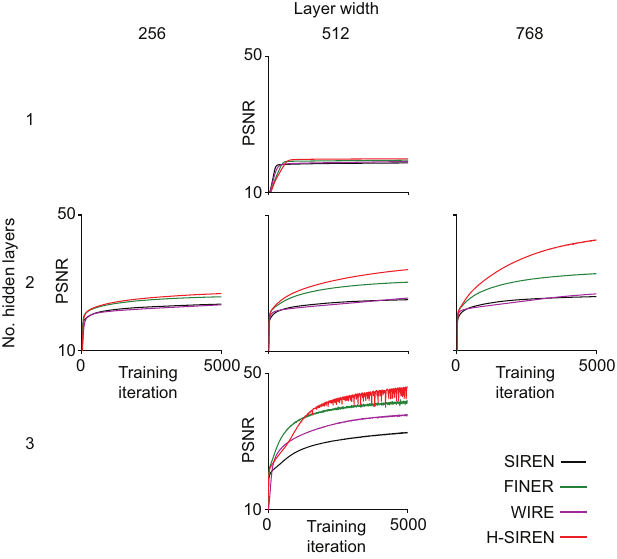}
	\caption{Comparison between different combinations of network width and depth for 2D image fitting task for image 0 in the "natural" data set. H-SIREN benefits more from layer width than other INRs.}
	\label{fig:fitimageothers}
\end{figure}

\paragraph{Additional results} We have experimented with a series of other network width and depth combinations. We plot out the PSNR values for the first image within the data set for these different combinations in Fig. \ref{fig:fitimageothers}. It can be observed that whilst all networks benefit from increasing layer depth, H-SIREN benefits significantly more from larger layer width than other INRs.

\section{Video fitting and video super-resolution}
\label{sec:appvideo}
\subsection{Implementation details}
\paragraph{Dataset} We use three different videos for the task, namely the "cat", "bike", and "bush" videos. The cat and bike videos are acquired from the publicly-available repository shared by Sitzmann et al. \cite{sitzmann2020implicit} with MIT licence, within which the cat video was originally acquired from \url{https://www.pexels.com/video/
the-full-facial-features-of-a-pet-cat-3040808/} with a free licence, and the bike video was originally acquired from scikit-video package \url{https://www.scikit-video.org/stable/datasets.html} with BSD licence. We collect and process the bush video ourselves. The cat video contains 300 frames, with each frame of resolution $512\times512$. The bike video contains 250 frames, with each frame of resolution $272\times640$. The bush video contains 300 frames, with each frame of resolution $600\times600$. 
\paragraph{Video fitting task} We use the whole cat and bike videos for the video fitting task. For the bush video, each frame is cropped to resolution $300\times300$ by only retaining its central part. Training is performed with Adam optimizer \cite{kingma2014adam} for 150 epochs, with batch size equal to the number of pixels in each frame, i.e., 262144 for the cat video, 174080 for the bike video, 90000 for the bush video. MSE loss is used throughout, and the learning rate is kept at $1\times10^{-4}$ for all three videos for all networks. All random seeds are initialized as 1 for all experiments. All the networks contain three hidden layers with layer width 1024. All inputs and outputs are normalized to have minimum -1 and maximum 1.
\paragraph{Video super-resolution task} We use the whole cat and bush videos for the video super-resolution task. For training, we sub-sample the videos at each frame by retaining every next pixel in both spatial directions, leading to a cat video of $256\times256$ pixels per frame, and a bush video of $300\times300$ pixels per frame. The networks are then trained on the sub-sampled data in the same way they are trained in the video fitting task. Unlike the video fitting task, after each training epoch, we evaluate the trained networks on the full-resolution video, and report the results with lowest mean-squared loss during training process for all the networks. All random seeds are initialized as 1 for all experiments.

\subsection{Additional results}
\begin{figure}
	\centering
	\includegraphics[]{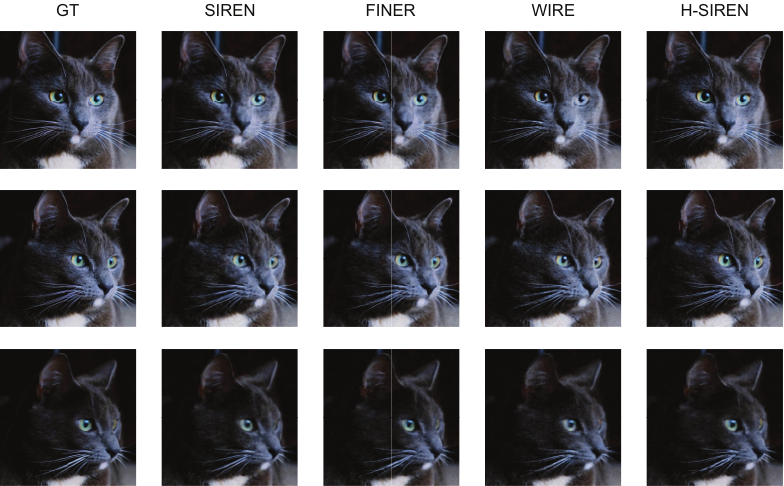}
	\caption{Comparison between different networks for super-resolution of the "cat" video. From top to bottom row: frame 10, 200, 290.}
	\label{fig:videosr}
\end{figure}

We plot out a few frames for the cat video in the video super-resolution task in Fig. \ref{fig:videosr} for the readers' reference.

\section{Neural radiance field}
\label{sec:appnerf}
\paragraph{Dataset} We take half of "blender" data set that is used in the original NeRF paper \cite{mildenhall2021nerf} that is publicly available with MIT licence, in particular the "drums", "mic", "lego", and "ship" scenes. The original train-eval-test split is used. All the training and tests are performed on the $400\times400$ images resampled from the original $800\times800$ images using pixel-area relation.
\paragraph{Implementation} We adapt from the implementation by Lin \cite{lin2020nerfpytorch} publicly available with MIT licence, which is a re-implementation of the original NeRF code in Pytorch. We changed the number of hidden layers in the original NeRF architecture from 8 to 3, and removed the skip connection. We removed the positional encoding for the input location and viewing direction except for the PE-ReLU case. To accommodate this change, we pass the input viewing direction through one network layer before concatenating it to the feature vector obtained from passing the input location through the network layers. We also removed the sigmoid activation before the RGB color output except for the PE-ReLU case, and replaced it with a simple scaling $x\rightarrow(x+1)/2$ for both the RGB color output and the opacity output. The layer width is set to be 362, which gives effective layer width 256 for WIRE. Other settings in NeRF are kept the same as the original implementation. We follow the hyperparameter choice of the authors for respective INRs: for FINER, we set $k=\frac{1}{\sqrt{3}}$; for WIRE, we set $\omega_0=\sigma_0=40$.
\paragraph{Training} We train the networks for 50000 iterations. For PE-ReLU, SIREN, FINER, and H-SIREN, the learning rate evolves from $10^{-4}$ to $10^{-5}$ during the training process following an exponential decay. For WIRE, the learning rate starts at $4\times10^{-4}$ following the choice of the original authors, and evolves to $4\times10^{-5}$ during the training process following an exponential decay. MSE loss in rendered color is used following the original NeRF paper. All random seeds are initialized as 1 for all experiments.

\section{Graph neural network-based fluid flow simulation}
\label{sec:appgnn}
\paragraph{Dataset} We use the flow around cylinder and flow around airfoil data sets from our earlier work \cite{gao2024finite}. The flow around cylinder data is simulated at Reynolds number $Re=200$, whilst we choose the flow around airfoil data set at $Re=2500$. We split both data sets such that 1025 continuous time steps are available for training, leading to a total of 1024 training samples with forward Euler time stepping. The test data set for the flow around cylinder contains 5465 continuous time steps, leading to a test set of size 5464 for 1-step prediction, and a test set to size 5454 for 10-step prediction. The test data set for the flow around airfoil contains 3466 continuous time steps, leading to a test set of size 3465 for 1-step prediction, and a test set to size 3455 for 10-step prediction. 
\paragraph{Implementation} We implement the MeshGraphNet \cite{pfaff2020learning} architecture. We reduce the layer depth to 5 since the data set is small, but kept the layer width of 128. All inputs and outputs, with the exception of the boundary feature one-hot vector, are normalized to have minimum -1 and maximum 1. For the test of different activation functions, we replace the activation function in all the multi-layer perceptrons in the MeshGraphNet, including the node and encoders, the node decoder, and the node and edge processors, with the activation function being tested. We increase the layer width of WIRE from 128 to 159 to approximately match the total number of parameters. For FINER, we select $k=1$ since it performs the best among all tested possible $k$ values. For WIRE, we directly use the default values of $\omega_0=30$ and $\sigma_0=10$ provided in the open-source code from the authors.
\paragraph{Training} We train the networks for 50 epochs which contains 10 warmup epochs at the start. Batch size is fixed at 4. The maximum learning rate is $10^{-4}$ and minimum learning rate is $10^{-6}$. We adopt the adaptive smooth-L1 loss and the merged cosine-exponential learning rate scheme in \cite{gao2024finite} to stabilize the training process. We do not adopt any training noise scheme. All random seeds are initialized as 1 for all experiments.

\section{Frequency bias tuning}
\label{sec:freqTune}
It is possible to further relax $f(x)$ in Eq. \ref{eq:sinh} by making it tunable, leading to the eventual form of the activation
\begin{equation}
	\begin{aligned}
		{x}^{l-1}=&{W}^l{z}^{l-1}+{b}^l\\
		{z}^{l}=&\sin(\omega_0\sinh(rx))
	\end{aligned}
\end{equation}
where the parameter $r=2$ as default. A lower $r$ will lead the network to bias towards lower frequencies, resulting in a smoother fitted profile, whilst a higher $r$ will also cover higher frequencies, leading to the network capturing more details. For some application scenarios, tuning the value of $r$ could provide a boost to the performance (cf. Table \ref{tab:varR}). We observed, however, that whilst the value of $r$ can theoretically be learned through backpropagation along with the other network parameters, it learns extremely slowly and needs to be manually designated at initialization in practice. We therefore report the results with the default value of $r=2$ in all the experiments presented in Sec. \ref{sec:exp}.
 
\begin{table}
	\caption{Results for various experiments with H-SIREN at different $r$ values. Error for GNN prediction are reported in RMSE, scaled by $\times10^{-4}$ for 1-step error and $\times10^{-5}$ for 10-step error.}
	\label{tab:varR}
	\centering
	\begin{tabular}{llllll}
		\toprule
		Image: \#3           &&&&&        \\
		\cmidrule(r){1-1}
		$r$ value & $1$ & $1.5$ & $2$ & $2.5$ & $3$ \\
		\midrule
		PSNR & $25.51\pm0.18$ & $30.37\pm0.10$ & $35.42\pm0.37$ & $37.923\pm0.84$ & $\mathbf{38.26\pm1.02}$ \\
		SSIM & $0.796\pm0.006$ & $0.921\pm0.004$ & $0.963\pm0.004$ & $0.976\pm0.005$ &  $\mathbf{0.976\pm0.007}$ \\
		\midrule
		Video: Bush           &&&&&        \\
		\cmidrule(r){1-1}
		$r$ value & $1$ & $1.5$ & $2$ & $2.5$ & $3$ \\
		\midrule
		PSNR & $19.14\pm0.64$ & $20.01\pm0.49$ & $20.31\pm0.46$ & $20.55\pm0.42$ & $\mathbf{20.57\pm0.36}$ \\
		SSIM & $0.501\pm0.041$ & $0.555\pm0.029$ & $0.568\pm0.025$ & $\mathbf{0.577\pm0.023}$ &  $0.575\pm0.022$ \\
		\midrule
		Scene: Lego           &&&&&        \\
		\cmidrule(r){1-1}
		$r$ value & $1$ & $1.5$ & $2$ & $2.5$ & $3$ \\
		\midrule
		PSNR & $26.32\pm1.26$ & $26.62\pm1.38$ & $\mathbf{26.75\pm1.35}$ & $26.56\pm1.32$ & $26.25\pm1.25$ \\
		SSIM & $0.885\pm0.018$ & $0.895\pm0.018$ & $\mathbf{0.900\pm0.017}$ & $0.897\pm0.016$ &  $0.893\pm0.015$ \\
		\midrule
		GNN: airfoil           &&&&&        \\
		\cmidrule(r){1-1}
		$r$ value & $1$ & $1.5$ & $2$ & $2.5$ & $3$ \\
		\midrule
		1-step & $21.7\pm9.01$ & $20.2\pm9.50$ & $20.0\pm9.96$ & $19.8\pm9.97$ &  $\mathbf{19.1\pm9.84}$ \\
		10-step & $15.8\pm5.7$ & $14.2\pm5.7$ & $13.7\pm6.1$ & $13.4\pm6.1$ &  $\mathbf{13.0\pm6.5}$ \\
		\bottomrule
	\end{tabular}
\end{table}

\end{document}